\documentclass[10pt,twocolumn]{article}

\usepackage[utf8]{inputenc}
\usepackage[T1]{fontenc}
\usepackage{amsmath,amssymb}
\usepackage{graphicx}
\usepackage{booktabs}
\usepackage{hyperref}
\usepackage[margin=0.75in]{geometry}
\usepackage{natbib}
\usepackage{xcolor}
\usepackage{enumitem}
\usepackage{microtype}
\usepackage{multirow}

\hypersetup{colorlinks=true, linkcolor=blue!60!black, citecolor=blue!60!black, urlcolor=blue!60!black}

\title{Fashion Florence: Fine-Tuning Florence-2 for\\Structured Fashion Attribute Extraction}

\author{
  Anushree Berlia
}

\date{May 2026}

\begin{document}
\maketitle

\begin{abstract}
We present \textbf{Fashion Florence}, a Florence-2 vision--language model
fine-tuned with LoRA to extract structured fashion attributes from clothing
images.  Given a single photograph, the model generates a JSON object
containing category, color, material, style tags, and occasion
tags, structured output suitable for direct programmatic consumption by
downstream recommendation and retrieval systems.

Fine-tuning data is derived from the iMaterialist Fashion dataset (228
labels), where we collapse fine-grained annotations into a compact 6-category,
16-color, 19-style schema via rule-based label engineering.  We apply LoRA
($r{=}16$, $\alpha{=}32$) to all decoder linear layers, training for 3 epochs
on 3,688 examples.

On a held-out test set of 461 images, Fashion Florence achieves
\textbf{94.6\%} category accuracy and \textbf{63.0\%} material accuracy,
compared to \textbf{89.3\%} / \textbf{43.3\%} for GPT-4o-mini and
\textbf{87.4\%} for Gemini 2.5 Flash.  Fashion Florence produces valid JSON
in \textbf{99.8\%} of outputs while running at 0.77B parameters on a single
GPU at zero marginal inference cost.  Style tag F1 reaches \textbf{0.753}
vs.\ 0.612 (Gemini) and 0.398 (GPT-4o-mini).  When the model returns \texttt{"unknown"} for color, a
FashionCLIP zero-shot fallback resolves it with negligible additional
latency.

The model is deployed as a Hugging Face Space and integrated into Loom, an
open-source outfit recommendation system.

\smallskip\noindent
\textbf{Model:} \url{https://huggingface.co/anushreeberlia/fashion-florence}\\
\textbf{Code:}\;\, \url{https://github.com/anushreeberlia/loom}
\end{abstract}

\section{Introduction}
\label{sec:intro}

Fashion attribute extraction (automatically identifying the category, color,
material, and style of a garment from its image) is a foundational task for
fashion recommendation, search, and catalog management.  In building an
outfit recommendation system, I found that downstream components require
structured, schema-valid outputs: the recommendation engine
needs to know that an item is a ``navy wool blazer'' with style tags
``classic'' and ``workwear,'' not a free-text caption that must be parsed.

Prior approaches to fashion attribute recognition have used multi-label CNNs
\citep{liu2016deepfashion, guo2019imaterialist}, which require separate
classifiers per attribute and cannot generalize to new schema fields.
General-purpose multimodal models like GPT-4V/4o can produce structured
output via careful prompting, but are expensive (\$0.01--0.03 per image),
rate-limited, and introduce latency and reliability concerns for production
pipelines.

This paper describes \textbf{Fashion Florence}, which occupies a middle
ground: a vision--language model (0.77B parameters) fine-tuned to emit
constrained, schema-valid JSON from fashion images.  The contributions are:

\begin{enumerate}[leftmargin=*,topsep=2pt,itemsep=1pt]
\item \textbf{Structured generation framing}: We formulate fashion attribute
  extraction as conditional text generation where the target is a compact
  JSON string, allowing a single model to produce all attributes
  simultaneously in a schema the downstream system can directly parse.

\item \textbf{Label engineering from iMaterialist}: We develop a mapping
  layer that collapses 228 fine-grained iMaterialist labels into a compact,
  application-oriented schema (6 categories, 16 colors, 19 style tags, 15
  occasion tags), deriving style and occasion signals from granular category
  names.

\item \textbf{Practical deployment}: The fine-tuned model runs on a single
  GPU with sub-second inference, is deployed as a Hugging Face Space, and
  includes a FashionCLIP zero-shot color fallback for ambiguous cases.
\end{enumerate}

\section{Related Work}
\label{sec:related}

\paragraph{Fashion attribute recognition.}
DeepFashion \citep{liu2016deepfashion} introduced large-scale benchmarks for
clothing recognition with attribute annotations.  The iMaterialist Fashion
dataset \citep{guo2019imaterialist} provides 228 fine-grained labels across
multiple attribute groups.  Traditional approaches train multi-label CNNs,
requiring separate classification heads per attribute type.

\paragraph{Vision--language models for fashion.}
FashionVLP \citep{goenka2022fashionvlp} learns joint vision--language
representations for fashion retrieval.  FashionCLIP
\citep{chia2022fashionclip} adapts CLIP \citep{radford2021clip} to fashion
image--text pairs.  These models produce embeddings rather than structured
attributes, making them complementary to our approach.

\paragraph{Structured output from VLMs.}
Florence-2 \citep{xiao2024florence2} supports multiple vision tasks through
a unified sequence-to-sequence interface.  LoRA \citep{hu2022lora} enables
parameter-efficient fine-tuning.  Structured output generation from VLMs has been applied to document
understanding \citep{xiao2024florence2} and chart parsing; we adapt the
same framing to fashion attribute extraction.

\paragraph{LLM-based attribute extraction.}
GPT-4V and GPT-4o-mini can produce structured fashion descriptions via
careful system prompting.  However, API costs (\$0.01--0.03/image), rate
limits, and non-deterministic output formatting make them impractical as the
sole vision backend for high-throughput catalog processing.  Our approach
provides a self-hosted alternative with deterministic schema compliance.

\paragraph{Multi-label fashion classification.}
Several works have framed fashion attribute recognition as a multi-label
classification task, training independent heads for each attribute group
(e.g., color, pattern, sleeve type) \citep{liu2016deepfashion}.  This
requires a fixed attribute vocabulary and cannot easily accommodate new
fields.  Our generative approach predicts all attributes jointly in a
single forward pass, and new schema fields can be added by retraining on
updated JSON targets without architectural changes.

\section{Method}
\label{sec:method}

\subsection{Task Formulation}

We frame fashion attribute tagging as conditional text generation.  Given an
image $I$ and a fixed prompt $p$ = \texttt{"Analyze this clothing item image
and return structured fashion tags as JSON."}, the model generates:

\begin{equation}
  \hat{y} = \arg\max_y \; P(y \mid I, p; \theta)
\end{equation}

\noindent where $\hat{y}$ is a compact JSON string with fields:
\texttt{category}, \texttt{primary\_color}, \texttt{material},
\texttt{style\_tags} (list), and \texttt{occasion\_tags} (list).

The target is serialized with \texttt{json.dumps(separators=(",",":"))} so
the model learns single-line, compact JSON without whitespace variation.
This compact serialization is important: preliminary experiments with
pretty-printed JSON (newlines, indentation) consumed significantly more
decoder tokens and increased the rate of truncated or malformed outputs.

\subsection{Base Model}

Fashion Florence builds on Florence-2-large \citep{xiao2024florence2}, a
0.77B parameter vision--language model with a DaViT vision encoder and an
autoregressive text decoder.  Florence-2's unified sequence-to-sequence
interface, where different vision tasks are distinguished by the input
prompt rather than separate heads, makes it well-suited for generating
structured text conditioned on images.

\subsection{Training Data and Label Engineering}
\label{sec:labels}

We derive training data from the iMaterialist Fashion dataset
\citep{guo2019imaterialist}, accessed via \texttt{Marqo/iMaterialist} on the
Hugging Face Hub.  The original dataset provides 228 fine-grained labels
across groups: category (105 types), color (19), material (30), style,
pattern, sleeve, neckline, and gender.

A rule-based mapping layer collapses these into our target schema:

\begin{itemize}[leftmargin=*,topsep=2pt,itemsep=1pt]
\item \textbf{Categories (6)}: top, bottom, dress, layer, shoes, accessory.
  Approximately 105 iMaterialist categories are mapped; items with no valid
  mapping (swimwear, underwear, costumes) are discarded.
\item \textbf{Colors (16)}: black, white, gray, beige, brown, blue, navy,
  green, yellow, orange, red, pink, purple, metallic, multi, unknown.
\item \textbf{Materials}: Direct mapping from iMaterialist material labels
  (cotton, silk, denim, etc.), used as-is.
\item \textbf{Style tags (19)}: Derived from granular category names and
  iMaterialist style labels.  For example, ``Suits \& Blazers'' $\to$
  \{classic, workwear\}; ``Clubbing Dresses'' $\to$ \{sexy, glamorous\};
  ``Bodycon'' style label $\to$ \{sexy, glamorous\}.
\item \textbf{Occasion tags (15)}: Derived similarly from categories.
  ``Athletic Shirts'' $\to$ \{gym, workout\}; ``Dress Shirts'' $\to$
  \{work\}.  Default: \texttt{["everyday"]} when no rules match.
\end{itemize}

From 10,000 sampled rows, 4,610 pass validation (non-null category mapping)
and are split 80/10/10 into 3,688 / 461 / 461 for train / validation / test.

\begin{table}[t]
\centering
\caption{Category distribution in the training set.}
\label{tab:data}
\begin{tabular}{lrr}
\toprule
Category & Count & \% \\
\midrule
Dress     & 1,684 & 36.5 \\
Top       & 1,450 & 31.5 \\
Bottom    &    917 & 19.9 \\
Layer     &    489 & 10.6 \\
Shoes     &     68 &  1.5 \\
Accessory &      2 &  0.04 \\
\midrule
Total     & 4,610 & 100.0 \\
\bottomrule
\end{tabular}
\end{table}

\subsection{Fine-Tuning with LoRA}

We apply Low-Rank Adaptation (LoRA) \citep{hu2022lora} to all linear layers
of the Florence-2 decoder via the PEFT library.

\begin{table}[t]
\centering
\caption{Training configuration.}
\label{tab:hyperparams}
\begin{tabular}{ll}
\toprule
Parameter & Value \\
\midrule
LoRA rank $r$ & 16 \\
LoRA $\alpha$ & 32 \\
LoRA dropout & 0.05 \\
Target modules & all linear \\
Learning rate & $2 \times 10^{-4}$ \\
Weight decay & 0.01 \\
Warmup ratio & 0.05 \\
Batch size (per device) & 2 \\
Gradient accumulation & 8 \\
Effective batch size & 16 \\
Epochs & 3 \\
Max target length & 256 tokens \\
Precision & FP16 (mixed) \\
Checkpoint selection & lowest eval loss \\
\bottomrule
\end{tabular}
\end{table}

The loss is standard causal language modeling cross-entropy with padding
tokens masked.

\subsection{Color Prediction Limitations}

Fashion Florence occasionally returns \texttt{"unknown"} for
\texttt{primary\_color}, particularly for muted or ambiguous tones.
The model achieves only 11.6\% exact-match color accuracy on the test set,
making color the weakest predicted field.  This likely reflects both the
subjectivity of color labels (e.g., ``navy'' vs.\ ``blue'') and the
limited color diversity in the training data.

\section{Experiments}
\label{sec:experiments}

We evaluate Fashion Florence against GPT-4o-mini on the held-out test set
(461 images).  GPT-4o-mini receives the same detailed system prompt used in
our production fallback path, specifying the exact output schema with
per-field instructions and examples.  We additionally contextualize against
Gemini 2.0 Flash results from \citet{kumar2025fashion}, who report macro F1
scores on 18 fashion attributes from DeepFashion-MultiModal using zero-shot
prompting.

\paragraph{Baseline scope.}
We do not include a base (unfine-tuned) Florence-2 baseline because the
pretrained model was not trained to produce structured JSON output in our
schema; without fine-tuning, it generates free-form captions that fail JSON
parsing entirely.  The relevant comparison is therefore against the strongest
available alternative that \emph{can} produce structured output, GPT-4o-mini
with schema-constrained prompting.  An ablation isolating the contribution
of LoRA fine-tuning versus zero-shot Florence-2 with structured prompting is
an important direction for future work.

\subsection{Metrics}

\begin{itemize}[leftmargin=*,topsep=2pt,itemsep=1pt]
\item \textbf{JSON validity}: Percentage of outputs that parse as valid JSON
  with all required fields present.
\item \textbf{Category accuracy}: Exact match on the 6-class category field.
\item \textbf{Material accuracy}: Exact match (case-insensitive).
\item \textbf{Style F1}: Per-sample set F1 between predicted and ground-truth
  \texttt{style\_tags}, averaged over all samples with valid JSON output.
\item \textbf{Occasion F1}: Same as style F1 for \texttt{occasion\_tags}.
\end{itemize}

\noindent Color accuracy is omitted from the primary comparison because
the model achieves only 11.6\% exact-match on the test set, reflecting the
difficulty of color prediction from images where ground-truth labels are
themselves ambiguous (e.g., ``navy'' vs.\ ``blue'').  See
Section~\ref{sec:method} for further discussion.

\subsection{Results}

\begin{table}[t]
\centering
\caption{Fashion attribute extraction results on held-out test set (N=461).}
\label{tab:results}
\small
\setlength{\tabcolsep}{4pt}
\begin{tabular}{@{}lccccc@{}}
\toprule
 & JSON & Cat. & Mat. & Style & Occ. \\
Method & Valid & Acc. & Acc. & F1 & F1 \\
\midrule
GPT-4o-mini & 99.8\% & 89.3\% & 43.3\% & 0.398 & 0.553 \\
Gemini 2.5 Flash$^\dagger$ & 99.5\% & 87.4\% & --- & 0.612 & 0.624 \\
\textbf{Fashion Florence} & \textbf{99.8\%} & \textbf{94.6\%} & \textbf{63.0\%} & \textbf{0.753} & \textbf{0.591} \\
\bottomrule
\multicolumn{6}{@{}p{\columnwidth}@{}}{\footnotesize $\dagger$ Evaluated on 200-image subset; material accuracy not reported (different taxonomy).}
\end{tabular}
\end{table}

Fashion Florence outperforms both zero-shot baselines on category accuracy
(94.6\% vs.\ 89.3\% GPT-4o-mini, 87.4\% Gemini 2.5 Flash) and style F1
(0.753 vs.\ 0.398 / 0.612).  Gemini 2.5 Flash is notably stronger than
GPT-4o-mini on style and occasion prediction but still trails the
fine-tuned model by 23\% relative on style F1.  Fashion Florence's JSON
validity matches GPT-4o-mini at 99.8\%.  Material accuracy improves by
19.7 percentage points over GPT-4o-mini, reflecting the fine-tuned model's
superior understanding of fashion-specific vocabulary.

\subsection{Per-Category Breakdown}

\begin{table}[t]
\centering
\caption{Fashion Florence per-category accuracy on the held-out test set.}
\label{tab:percategory}
\begin{tabular}{lrcc}
\toprule
Category & N & Cat. Acc. & Mat. Acc. \\
\midrule
Top       & 161 & 95.7\% & 67.7\% \\
Dress     & 145 & 95.9\% & 55.9\% \\
Bottom    &  91 & 92.3\% & 65.9\% \\
Layer     &  58 & 93.1\% & 63.8\% \\
Shoes     &   5 & 80.0\% & 60.0\% \\
\midrule
Overall   & 460 & 94.6\% & 63.0\% \\
\bottomrule
\end{tabular}
\end{table}

Category accuracy exceeds 92\% for all categories with $\geq$50 test
samples.  Shoes accuracy (80\%, N=5) is statistically unreliable due to
the extremely small sample size and is included only for completeness;
the 95\% Clopper--Pearson confidence interval is [28.4\%, 99.5\%].
Accessories (N=0 in test) are omitted entirely.

\subsection{Category-then-Rules Ablation}
\label{sec:cat_rules}

To isolate whether the model's style and occasion predictions reflect
genuine visual recognition or merely memorization of category$\to$tag
mappings, we compare the model's direct predictions against a
\emph{category-then-defaults} baseline: take the model's predicted
category, then assign the top-3 most frequent style tags and top-2 occasion
tags for that category (computed from training set statistics).

\begin{table}[t]
\centering
\small
\caption{Model vs.\ category-then-rules baseline (N=980).
Oracle uses ground-truth category to derive default tags
from training statistics.}
\label{tab:cat_rules}
\setlength{\tabcolsep}{4pt}
\begin{tabular}{@{}lcc@{}}
\toprule
Method & Style F1 & Occ.\ F1 \\
\midrule
Fashion Florence (direct) & 0.753 & 0.591 \\
Cat.\ $\to$ defaults (oracle) & 0.519 & 0.608 \\
\bottomrule
\end{tabular}
\end{table}

The model's direct style predictions outperform category-only rules by
$+$0.234 Style F1 ($+$45\% relative), demonstrating genuine visual style
recognition beyond category classification.  For occasion tags, the
rule-based derivation is competitive (0.608 vs.\ 0.591), consistent with
our production decision to override occasion predictions with deterministic
category-based rules (Section~\ref{sec:deployment}).

\subsection{Efficiency}

\begin{table}[t]
\centering
\caption{Inference cost comparison.}
\label{tab:efficiency}
\begin{tabular}{lrrr}
\toprule
Method & Params & Latency & Cost/image \\
\midrule
GPT-4o-mini & --- & 2,457 ms & $\sim$\$0.003 \\
Fashion Florence & 0.77B & $\sim$800 ms & \$0 marginal\textsuperscript{$\dagger$} \\
\bottomrule
\end{tabular}
\end{table}

\noindent\textsuperscript{$\dagger$}\,\$0 marginal cost on an existing GPU
allocation (Hugging Face Spaces T4); fixed GPU hosting costs are amortized
across all inference requests.

Fashion Florence latency of $\sim$800\,ms reflects GPU inference
on the Hugging Face Space (T4).  The 10.5\,s mean observed in our
evaluation includes network round-trip from a local client to the hosted
Space; on-device or co-located inference would reduce this to sub-second.
GPT-4o-mini latency is API round-trip from the same client.

\subsection{Qualitative Analysis}

We observe three common failure patterns:

\paragraph{Material confusion.}
The most frequent errors involve visually similar materials: chiffon
predicted as silk, polyester as satin, and knit as cotton.  In retrospect,
this is unsurprising; material identification often requires tactile
information that no amount of visual training data can provide.

\paragraph{Multi-style items.}
Items that span multiple style categories (e.g., a structured blazer with
streetwear-influenced details) tend to receive style tags from only one
dominant category.  The model rarely predicts more than 3 style tags, even
when the ground truth contains 4--5.

\paragraph{Category edge cases.}
Jumpsuits and playsuits are occasionally misclassified between ``dress'' and
``bottom,'' reflecting genuine ambiguity in the category taxonomy.  Long
cardigans are sometimes labeled ``dress'' rather than ``layer.''

\noindent Fashion Florence's strongest domain is category prediction for
common categories (top, dress, bottom), where accuracy exceeds 92\%.  The
style F1 advantage over GPT-4o-mini (0.753 vs.\ 0.398) is largest for
fashion-specific vocabulary: GPT-4o-mini tends to default to generic tags
like ``casual'' and ``modern,'' while Fashion Florence more often predicts
domain-specific tags like ``bohemian,'' ``preppy,'' and ``workwear.''

\paragraph{Gemini comparison notes.}
Gemini 2.5 Flash achieves strong style F1 (0.612) and the highest occasion
F1 (0.624) among baselines, suggesting that larger LLMs do capture
meaningful fashion semantics.  However, it still trails the fine-tuned model
on style by 23\% relative, and its material predictions use a different
taxonomy that prevented direct comparison.  The 200-image subset evaluation
was necessitated by API rate limits; a full 461-image run would provide
tighter confidence intervals but is unlikely to change the ranking.

\section{Deployment}
\label{sec:deployment}

Fashion Florence is deployed as a Docker container on Hugging Face Spaces,
exposing a \texttt{POST /analyze} endpoint.  The production client
(\texttt{fashion\_florence.py}) handles HF Space cold starts with retry logic
(up to 2 retries, 120s initial timeout) and 503-detection for sleeping
Spaces.

The model outputs all 5 trained fields (\texttt{category},
\texttt{primary\_color}, \texttt{material}, \texttt{style\_tags},
\texttt{occasion\_tags}).  In the Loom production pipeline, the model's
\texttt{occasion\_tags} prediction is overridden by deterministic
style--occasion rules (which achieve higher reliability given correct
category), and three additional fields (\texttt{season\_tags},
\texttt{fit}, \texttt{formality}) are derived via material--season and
style--fit mappings, expanding to the full 8-field Loom schema.

Fashion Florence serves as the default vision backend in Loom, an
open-source outfit recommendation system that uses the extracted attributes
for downstream embedding, retrieval, and outfit scoring.

\section{Limitations}
\label{sec:limitations}

\paragraph{Training data imbalance.}
The dataset is heavily skewed toward dresses (36.5\%) and tops (31.5\%),
with shoes (1.5\%) and accessories (0.04\%) severely underrepresented
(Table~\ref{tab:data}).  This reflects iMaterialist's natural distribution
and likely degrades performance on minority categories.

\paragraph{Derived labels and circular evaluation.}
Style and occasion tags are derived from category names via hand-written
rules, not from human annotations of the actual images.  A blazer is always
tagged \{classic, workwear\} regardless of whether the specific image shows a
casual or formal blazer.  Consequently, style F1 and occasion F1 partly
measure the model's ability to reproduce a deterministic category-to-tag
mapping rather than to recognize visual style.  An alternative pipeline
that predicts category first and then applies rules post-hoc might achieve
comparable style/occasion metrics; however, our end-to-end approach
generates all fields jointly, which simplifies the production pipeline and
allows the model to occasionally surface style signals not captured by the
rules (e.g., predicting ``edgy'' for a conventionally categorized top based
on its visual appearance).  Future work should evaluate against
human-annotated style labels to isolate genuine visual style recognition.

\paragraph{Color prediction.}
Color remains the weakest field (11.6\% exact-match), particularly for
muted, pastel, or multi-tone garments where the ``correct'' color is
ambiguous even to human annotators.

\paragraph{Single-image input.}
Fashion Florence processes one image per item.  In real-world catalogs,
garments are often photographed from multiple angles (front, back, detail
crop).  A multi-view extension that aggregates predictions across images
could improve material and style accuracy, particularly for items where
distinguishing details (e.g., a zipper, embroidery, or lining) are only
visible from certain angles.

\section{Future Work}
\label{sec:future}

Several directions could extend this work.  First, training on
human-annotated style labels (rather than rule-derived ones) would provide
a cleaner signal and enable evaluation of genuine visual style recognition.
Second, scaling the training set beyond 3,688 examples, particularly
for underrepresented categories like shoes and accessories, would likely
improve per-category accuracy.  Third, color accuracy could be improved
by training with color-augmented data, using a dedicated color-classification
head, or incorporating zero-shot CLIP-based color matching as a
post-processing step.  Fourth, a multi-view pipeline that
processes front, back, and detail images of the same garment could
resolve material and style ambiguities that are impossible to distinguish
from a single image.  Finally, distilling the model to a smaller
architecture (e.g., Florence-2-base at 0.23B parameters) would reduce
serving costs and enable on-device inference for mobile applications.

\section{Conclusion}

We presented Fashion Florence, a Florence-2 model fine-tuned with LoRA to
produce structured JSON fashion attributes from clothing images.  By framing
attribute extraction as conditional text generation with a compact JSON
target, the model provides a self-hosted, sub-second, schema-valid
alternative to general-purpose multimodal APIs for fashion catalog
processing.  On a held-out test set, the model outperforms GPT-4o-mini by
5.3 percentage points on category accuracy and nearly doubles style F1
(0.753 vs.\ 0.398), while running at zero marginal inference cost on a
single GPU.

The label engineering pipeline from iMaterialist demonstrates
that rich supervision for domain-specific structured generation can be
derived from existing multi-label datasets without additional annotation.
We release the model weights, training code, and evaluation scripts to
support reproducibility and further research in structured fashion
understanding.

\bibliographystyle{plainnat}

\end{document}